\documentclass[letterpaper, 10 pt, conference]{ieeeconf}  

\IEEEoverridecommandlockouts                              

\usepackage{graphicx}
\usepackage{setspace}
\usepackage{dsfont}

\usepackage{amsmath,amssymb, amsfonts, bm} 
\usepackage{url}
\usepackage{doi}
\usepackage{tabularx}
\usepackage[font={small}]{caption}
\usepackage{subcaption}
\usepackage{dblfloatfix}
\usepackage[linesnumbered,ruled,vlined]{algorithm2e}
\usepackage{multirow}
\usepackage{color}
\usepackage{cite}
\usepackage{hyperref}   
\usepackage{booktabs} 
\usepackage{array}    
\newcolumntype{C}[1]{>{\centering\arraybackslash}m{#1}}
\newcolumntype{L}[1]{>{\raggedright\arraybackslash}m{#1}}
\newcolumntype{R}[1]{>{\raggedleft\arraybackslash}m{#1}}

\usepackage[linesnumbered,ruled,vlined]{algorithm2e}
\usepackage{amssymb}
\usepackage{xcolor} 

\usepackage{ifthen}
\newboolean{includeFigures}
\setboolean{includeFigures}{true}

\usepackage[font=small,skip=1pt]{caption}

\definecolor{drakgreen}{rgb}{0.0, 0.5, 0.0}  

\usepackage{amsthm}
\theoremstyle{definition} 
\newtheoremstyle{italdefinition} 
{1pt} 
{1pt} 
{} 
{10pt} 
{\itshape} 
{:} 
{3pt} 
{} 
\theoremstyle{italdefinition}

\usepackage{hyperref}
\hypersetup{
    colorlinks=true,
    linkcolor=blue,
    filecolor=magenta,      
    urlcolor=blue,
    pdftitle={Manipulation Robustness in TAMP},
    pdfpagemode=FullScreen,
    }

\author{Yifei Dong$^{1,*}$, Yan Zhang$^{2,*}$, Sylvain Calinon$^{2}$, Florian T. Pokorny$^{1}$
\thanks{$^{1}$The authors are with the division of Robotics, Perception and Learning, KTH Royal Institute of Technology, 10044 Stockholm, Sweden. 
$^{2}$The authors are with the Idiap Research Institute, CH-1920 Martigny,
Switzerland and also with the EPFL, 1015 Lausanne, Switzerland.
Authors with $^{*}$ contributed equally.
Funded by the European Commission under the Horizon Europe Framework Program project
SoftEnable, grant number 101070600, {\url{https://softenable.eu/}} and the State Secretariat for Education, Research and Innovation in Switzerland for participation in the European Commission’s Horizon Europe Program through the INTELLIMAN project (\url{https://intelliman-project.eu/}) and the SESTOSENSO project (\url{http://sestosenso.eu/}).
Contact: {\tt\small yifeid@kth.se, yan.zhang@idiap.ch}.
}}

\usepackage{graphicx}
\def\BibTeX{{\rm B\kern-.05em{\sc i\kern-.025em b}\kern-.08em
    T\kern-.1667em\lower.7ex\hbox{E}\kern-.125emX}}
\usepackage[font=small,labelfont=bf,tableposition=top]{caption}

\usepackage{blindtext}
\title{\LARGE \bf
Robustness-Aware Tool Selection and Manipulation Planning with Learned Energy-Informed Guidance
}

\begin{document}

\maketitle
\thispagestyle{empty}
\pagestyle{empty}
\begin{abstract}
Humans subconsciously choose robust ways of selecting and using tools, for example, choosing a ladle over a flat spatula to serve meatballs. However, robustness under external disturbances remains underexplored in robotic tool-use planning. This paper presents a robustness-aware method that jointly selects tools and plans contact-rich manipulation trajectories, explicitly optimizing for robustness against disturbances. At the core of our method is an energy-based robustness metric that guides the planner toward robust manipulation behaviors. We formulate a hierarchical optimization pipeline that first identifies a tool and configuration that optimizes robustness, and then plans a corresponding manipulation trajectory that maintains robustness throughout execution. We evaluate our method across three representative tool-use tasks. Simulation and real-world results demonstrate that our method consistently selects robust tools and generates disturbance-resilient manipulation plans.
\end{abstract}
    

\section{Introduction} \label{sec-intro}
Humans develop an intuition for robustness in tool use from an early age. Even with limited precision, babies learn to complete tasks under uncertainty, and adults subconsciously leverage experience to select and use tools in robust ways. As shown in Fig.~\ref{fig:teaser}, different tools can lead to significantly different outcomes under disturbance, highlighting the importance of robustness in both selection and use. Despite this, robustness in robotic tool use remains underexplored, as most prior work prioritizes task completion over disturbance resilience. Achieving human-like robustness in robots is challenging for two reasons. First, tools vary widely in geometry and physical properties, making manipulation outcomes highly sensitive to the combined configuration of tool, robot, and object. Second, robustness requires not only planning effective tool-use strategies. It also requires selecting the best tool from a diverse set, while accounting for complex contact dynamics in uncertain, contact-rich environments.

To address these challenges, we propose a unified method that jointly selects the most robust tool and plans a corresponding manipulation trajectory. The problem is formulated as a hierarchical optimization process: (i) selecting a tool and tool-object interaction that is the most robust, and (ii) planning a manipulation trajectory passing through the selected interaction configuration while maintaining the robustness throughout task execution with the best-found tool. Since robustness evaluation is central but computationally demanding, we introduce an energy-informed metric that provides efficient robustness guidance at both stages of optimization. To further reduce online costs, we train a neural network offline on a dataset of tool–object configurations paired with robustness scores, enabling fast inference during planning. We evaluate our method on three representative tasks: tape pulling, scissors hooking, and fish scooping. The tasks encompass rigid, articulated, and deformable objects. Experimental results demonstrate that our method effectively selects a tool and manipulation trajectory robust against random force disturbances.

\ifthenelse{\boolean{includeFigures}}{
\begin{figure}[t]
\centering
\includegraphics[width=1\linewidth]{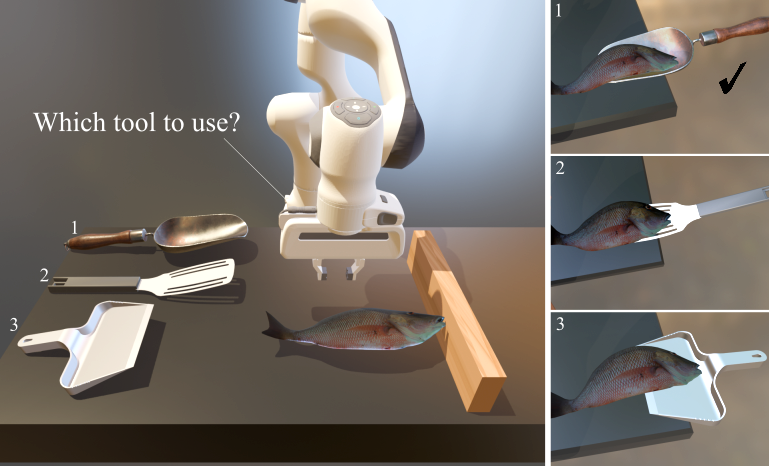}
\caption{How can the robot choose the best tool to scoop and lift a fish so that it does not fall out of the tool during manipulation? This work introduces a robustness-aware planner that selects and uses tools effectively under disturbances. 
}
\label{fig:teaser}
\end{figure}
}{}

In summary, our contributions include:
(i) a robustness-aware optimization method that jointly selects tools and plans contact-rich manipulation trajectories, prioritizing tools and configurations robust to disturbances,
(ii) a learned energy-informed robustness metric derived from caging analysis, which enables efficient online robustness guidance, and
(iii) simulation and real-world experiments demonstrating improved robustness and reliability compared to a baseline method.

\section{Related work}
\subsection{Robust Manipulation}
Robust manipulation aims to ensure consistent and reliable task execution in the presence of uncertainty, whether from noisy sensing, unmodeled dynamics, or unpredictable environmental interaction~\cite{wang2020feature, bhatt2022surprisingly}. Mason highlights the importance of robustness in tasks such as crank turning and peg-in-hole insertion, which succeed only if the system tolerates errors in object pose or robot motion~\cite{mason2018toward}. Common sources of uncertainty include geometric ambiguity due to limited perception~\cite{bohg2013data, daniels2023grasping} and dynamic uncertainty from contact interactions~\cite{andrychowicz2020learning, jankowski2024planning}. Traditional methods focus on prehensile manipulation, using analytic grasp metrics~\cite{pollard1996synthesizing, roa2015grasp} or learning-based estimators~\cite{saxena2008learning, mahler2017dex, mahler2018dex} to select robust grasps. Environmental contacts are often exploited by humans to reduce uncertainties during manipulation—for example, sliding an object along a surface to correct pose uncertainty before grasping it. Such uncertainty reduction strategies with action rather than sensing are formalized in funnels~\cite{mason1985mechanics} and applied in in-hand robust manipulation~\cite{bhatt2022surprisingly}.

In contrast, robustness in non-prehensile manipulation remains underexplored. 
Caging~\cite{kuperberg1990problems, rimon1999caging, rodriguez2012caging}  provides valuable insights into addressing these challenges. Caging is a strategy that prevents objects from escaping through geometric constraints without relying on force or form closure, allowing for the tolerance of geometric uncertainties. Energy-bounded caging and its variants~\cite{mahler2018synthesis, dong2024quasi, dong2024characterizing, shirizly2024selection, wang2024caging} extend caging by incorporating energy constraints on the object.
In this work, we integrate an energy-bounded caging metric as robustness guidance into a tool-use manipulation planner. This data-driven method improves online efficiency by offloading heavy computations of robustness labeling offline. It identifies robust tool-object configurations from an energy perspective and guides robust manipulation synthesis.

\subsection{Manipulation with Tools}
Tools can significantly extend the dexterity of robotic end-effectors, enabling more complicated manipulation tasks such as flipping pancakes \cite{Calinon13RAS}, rolling dough \cite{Calinon13IROS,shi2023robocook}, scooping tofu with a spoon~\cite{grannen2023learning}, pouring water with a box~\cite{seita2023toolflownet}, throwing with a ball~\cite{liu2024tube}, cutting food with a knife~\cite{heiden2021disect}, cleaning with a broom \cite{Silverio15IROS}, hammering nails \cite{Ti24TRO}, or performing brain surgery using neurosurgical instruments~\cite{he2025magnetically}. Robotic tool use often involves one or more tools, multiple sequential actions, and goals that consist of several sub-goals~\cite{qin2023robot}. In most existing robotic tool-use applications, tools are typically assumed to be either rigidly attached to the robot arm or grasped and held during use~\cite{suomalainen2022survey}.

Tool selection is defined as the process of identifying the most appropriate tool from a set of candidates to complete a specific task~\cite{qin2023robot}. This is a challenging problem due to the large combinatorial search space involved. Levihn et al.\cite{levihn2014using} propose an approach that reasons over physical constraints between objects to reduce computational complexity, while Xu et al.\cite{xu2023creative} utilize language models to guide tool selection or creation. Existing tool selection approaches often focus on semantic reasoning or functional feasibility, without modeling the geometry-dependent robustness of tool-object interactions. In contrast, we introduce a unified method that explicitly evaluates and optimizes for manipulation robustness during tool selection and motion planning.
Additionally, our research is closely related to recent studies on the co-design of robot behavior and tool morphology~\cite{li2023learning, liu2023learning, liu2024paperbot, dong2025cagecoopt}. Unlike tool design works that can tune or generate geometry to maximize performance, we operate under stricter constraints—selecting the best tool and usage strategy from a fixed set of available tools.

\ifthenelse{\boolean{includeFigures}}{
\begin{figure}[t]
    \centering
    \includegraphics[width=1\linewidth]{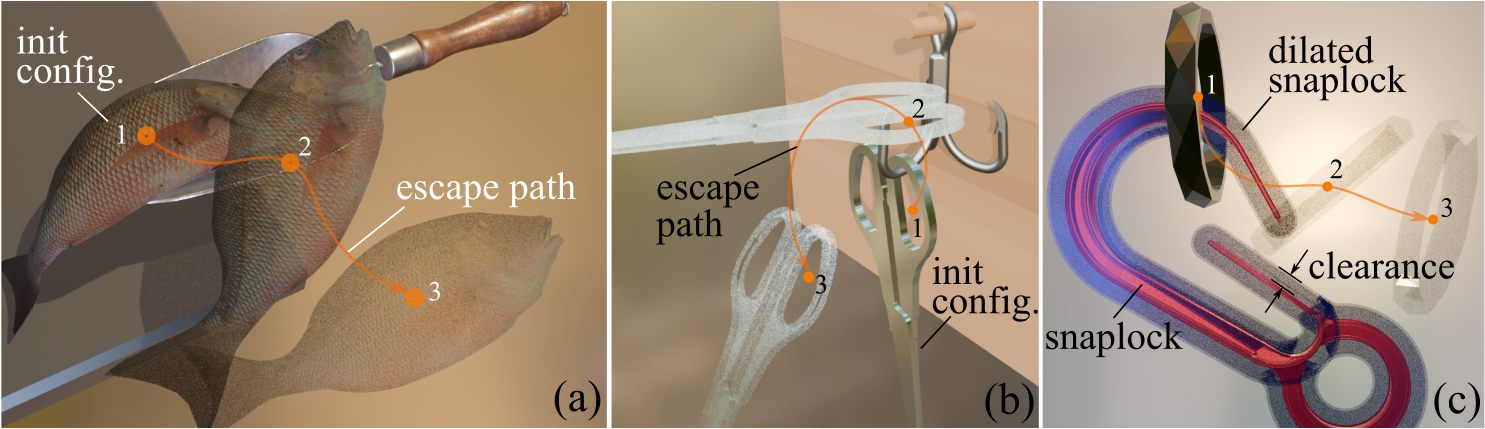}
    \caption{Manipulation robustness in three scenarios. Fish in a shovel (a) and scissors on a treble hook (b) remain secured unless large disturbances occur; keyframes (1–3) illustrate minimal-energy escape paths used to quantify robustness. The robustness of a keyring in a snaplock (c) can also be characterized by clearance, i.e., the minimal dilation needed to completely prevent escape.
    }
    \label{fig:ebc}
\end{figure}
}{}

\ifthenelse{\boolean{includeFigures}}{
\begin{figure*}[th]
    \centering
    \includegraphics[width=0.8\linewidth]{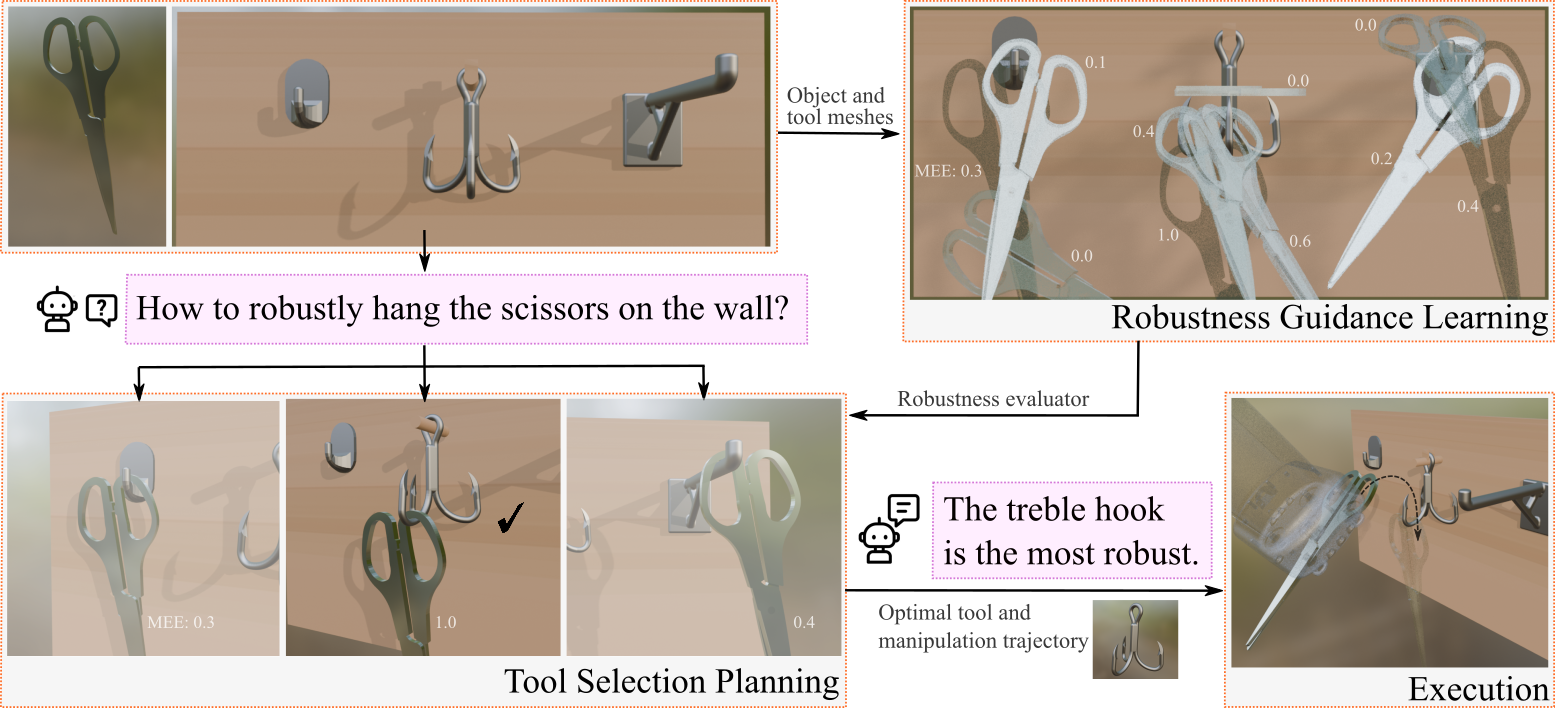}
    \caption{
    Overview of the robustness-aware tool selection and manipulation planning method. The pipeline first performs keyframe optimization to jointly select the most robust tool and tool–object configuration. The best-found configuration then serves as a waypoint for full trajectory optimization, which plans a robustness-preserving manipulation strategy. A neural network trained offline on simulated robustness scores enables efficient robustness evaluation throughout the planning process.
    }
    \label{fig:overview}
\end{figure*}
}{}

\section{Preliminary}
Here, we introduce an energy-informed metric derived from caging analysis, which guides robustness-aware manipulation planning~\cite{dong2024quasi}. Consider a rigid object's configuration space (C-space), denoted as \(\mathcal{S}_{\text{obj}}\), with \(\mathcal{S}_{\text{free}} \subset \mathcal{S}_{\text{obj}}\) representing its collision-free subset. 
In the presence of the force field, an object at configuration \( s_{\text{obj}} \in \mathcal{S}_{\text{free}} \) is associated with an energy value \( E(s_{\text{obj}}): \mathcal{S}_{\text{free}} \rightarrow \mathbb{R} \). The configuration's robustness is characterized by the \emph{Minimum Escape Energy} (MEE), denoted \({Q}_{\text{mee}}\). Intuitively, MEE~${Q}_{\text{mee}}$ measures how much additional energy the object needs to ``escape'' from its current configuration~$s_{\text{obj}}$. Take the fish in the shovel as an example (Fig.~\ref{fig:ebc}a). It has non-zero escape energy \({Q}_{\text{mee}} > 0\), meaning that it is secured by the shovel within a local subset of its configuration space and cannot escape unless external energy greater than \({Q}_{\text{mee}}\) is applied. In practice, we estimate the MEE by searching for the corresponding energy-efficient collision-free escape paths using sampling-based algorithms, such as BIT*~\cite{gammell2015batch}. For a formal definition of \({Q}_{\text{mee}}\) and details on how to estimate it using BIT*, we refer interested readers to \cite{dong2024quasi}.

We consider three forms of the energy function \( E \), depending on the task setting. (1) Under a gravitational field (Fig.~\ref{fig:ebc}b), the energy is given by \( {E(s_{\text{obj}}) = m_{\text{obj}}gz_{\text{obj}}} \), where \( m_{\text{obj}} \) is the object's mass, \( g \) is gravitational acceleration, and \( z_{\text{obj}} \) is its vertical position of its center of mass. (2) For planar pushing tasks, we consider an additional pushing energy field: \( {E(s_{\text{obj}}) = m_{\text{obj}}gz_{\text{obj}} + F_p\hat{v} \cdot (x_{\text{obj}}, y_{\text{obj}})} \), where \( F_p \) is the pushing force magnitude, \( \hat{v} \in \mathcal{S}^1 \) is a horizontal unit vector in the pushing direction, and \( (x_{\text{obj}}, y_{\text{obj}}) \) is the object’s planar position~\cite{mahler2018synthesis}. (3) For deformable objects (Fig.~\ref{fig:ebc}a), the energy includes an elastic term: \( E(s_{\text{obj}}) = m_{\text{obj}}gz_{\text{obj}} + \frac{1}{2}k\alpha_{\text{obj}}^2 \), where \( k \) is the stiffness coefficient and \( \alpha_{\text{obj}} \) is the deformation angle~\cite{dong2024quasi}. 

We use \textit{Partial Caging Clearance}~(PCC)\cite{welle2021partial} as a baseline to compare against MEE. Unlike MEE, which defines robustness from an energy perspective, PCC quantifies robustness based on the clearance between the object and the caging tools. PCC~$Q_{\text{pcc}}$ is defined as the minimal tool's geometric dilation offset~${\epsilon_{\text{min}}}$ required to prevent the object from escaping, provided no collision occurs (Fig.~\ref{fig:ebc}c). If such a collision-free offset ${\epsilon_{\text{min}}} \in [0, {\epsilon_{\text{col}}}]$ exists, $Q_{\text{pcc}}=-\epsilon_{\text{min}}$ and the object is ``partially caged'', denoted $q_{\text{pc}}=1$; otherwise it is not partially caged $q_{\text{pc}}=0$ and $Q_{\text{pcc}}$ is not defined. Practically, we approximate collision offset~$\epsilon_{\text{col}}$ using mesh distance and employ BIT* to detect escape paths, determining the partial cage status~$q_{\text{pc}}$ accordingly.

\section{Problem Statement and Formulation}\label{sec:problem-formulation}
Consider the problem of manipulating a target object \( o_{\text{obj}} \) from an initial configuration \( s_{\text{obj}}^{0} \) to a desired goal configuration \( s_{\text{obj}}^{G} \), using a tool selected from $N$ available ones \( \mathcal{O}_{\text{tool}} = \{o_i\}_{i=1}^{N} \). The objective is to determine the best tool \( o_{\text{tool}}^{*} \in \mathcal{O}_{\text{tool}} \) and corresponding tool trajectory \( \tau_{\text{tool}}^{*} = \{s_{\text{tool}}^{t}\}_{t=0}^{T} \) that maximize robustness~$Q$ while manipulating target object to its goal configuration $s_{\text{obj}}^{G}$:

\vspace{-0.3cm}

\begin{subequations}
    \begin{align}
    \min_{o_{\text{tool}}, \tau_{\text{tool}}} & \quad \|s_{\text{obj}}^{G} - s_{\text{obj}}^{T}\|_{2} - \beta\sum_{t=0}^{T} Q(s_{\text{obj}}^{t}, o_{\text{tool}}, s_{\text{tool}}^{t}), \\
    \text{s.t.} \quad & s_{\text{obj}}^{t+1}, s_{\text{tool}}^{t+1} = f(s_{\text{obj}}^{t}, o_{\text{tool}}, s_{\text{tool}}^{t}), \\
    \quad & g(s_{\text{obj}}^{t}, s_{\text{tool}}^{t}) \leq 0, \quad c(s_{\text{tool}}^{t}, r) \leq 0,
    \end{align}
\label{eq:main}
\end{subequations}

\vspace{-0.3cm}

\noindent where $s_{\text{obj}}^{T}$ denotes the final configuration of the target object at the terminal time step $T$. The metric~\( {Q}(s_{\text{obj}}^{t}, o_{\text{tool}}, s_{\text{tool}}^{t}) \) quantifies the robustness of the manipulation at each time step $t$, which depends on the selected tool $o_{\text{tool}}$, target object, and their configurations $s_{\text{tool}}^{t}$, $s_{\text{obj}}^{t}$. The constraint \( g(s^{t}_{\text{tool}}, s^{t}_{\text{obj}}) \) encodes the bounded free space $\mathcal{S}_{\text{free}}$ between the tool and target object. \( c(s^{t}_{\text{tool}}, r) \) captures the kinematic constraints while the robot $r$ manipulating the target object with tool~$o_{\text{tool}}$. $\beta$ is a tuning parameter that adjusts the balance between the terms of manipulation planning and robustness.
In the following, we present how we solve the optimization problem in Section~\ref{sec:planner}. Section~\ref{sec:metric_learning} details our data-driven method to learning the robustness metrics~$Q$, essential for robustness-aware guidance and computational efficiency of the optimization process. A method overview is illustrated in Fig.~\ref{fig:overview}.

\section{Robustness-Aware Tool Selection and Manipulation Planning}\label{sec:planner}
The problem of robustness-aware tool use is inherently hybrid, involving discrete decisions over tool choice and continuous optimization of manipulation trajectories. To make this mixed-integer optimization tractable, we adopt a hierarchical method to solve Eq.~\eqref{eq:main}. First, we perform keyframe optimization to identify the best tool~$o^*_{\text{tool}}$ and corresponding object and tool configurations~$s^*_{\text{obj}}$, $s^*_{\text{tool}}$ that maximize robustness ${Q}$. Given the best-found tool, we subsequently compute the manipulation trajectory~$\tau^*_{\text{tool}}$, aiming to maintain robustness while passing through the configurations~$s^*_{\text{obj}}$, $s^*_{\text{tool}}$ at the keyframe timestep~$T_k$.

\subsection{Keyframe Optimization}\label{subsec:keyframe_optimization}
We formulate the keyframe optimization problem as a mixed-integer program defined at a pre-specified keyframe timestep $T_k \in [0, T]$. The goal is to jointly select the most robust tool from the candidate set $\mathcal{O}_{\text{tool}} = \{o_i\}_{i=1}^{N}$ and determine the corresponding tool and object configurations $(s_{\text{tool}}, s_{\text{obj}})$ that maximizes the robustness metric $Q$:
\begin{subequations}
    \begin{align}
        \max_{o_{\text{tool}}, s_{\text{tool}}, s_{\text{obj}}} & \quad Q(s_{\text{tool}}, o_{\text{tool}}, s_{\text{obj}}),  \\
        \text{s.t.} \quad & g(s_{\text{tool}}, s_{\text{obj}}) \leq 0, \quad c(s_{\text{tool}}, r) \leq 0. \label{eq:keyframe_constr}
    \end{align}
\label{eq:keyframe_opt}
\end{subequations}

\vspace{-0.3cm}

\noindent This yields the best-found tool $o^*_{\text{tool}}$ and the corresponding configurations $(s^*_{\text{tool}}, s^*_{\text{obj}})$ at the keyframe timestep $T_k$. The resulting configurations serve as an intermediate waypoint that the robot aims to pass through during manipulation. Once established, the tool is used to guide the object toward its goal configuration $s^{\text{G}}_{\text{obj}}$, maintaining high robustness throughout the trajectory.

\subsection{Robust Tool Manipulation Planning}
Given the selected tool~$o^*_{\text{tool}}$ and keyframe configurations $(s^*_{\text{tool}}, s^*_{\text{obj}})$, we plan the manipulation trajectory~$\tau^*_{\text{tool}}$ that maintains robustness while guiding the object to its goal configuration. This is formulated as:
\begin{subequations}
    \begin{align}
        \min_{\tau_{\text{tool}}} & \quad \|s_{\text{obj}}^{G} - s_{\text{obj}}^{T}\|_{2} + \|s_{\text{tool}}^{*} - s_{\text{tool}}^{T_{k}}\|_{2} + \|s_{\text{obj}}^{*} - s_{\text{obj}}^{T_{k}}\|_{2}\notag \\
        \quad &  - \beta^{\prime}\sum_{t=0}^{T} {Q}(s_{\text{obj}}^{t}, o_{\text{tool}}^{*}, s_{\text{tool}}^{t}), \label{eq:tool_motion} \\
        \text{s.t.} \quad & s_{\text{obj}}^{t+1}, s_{\text{tool}}^{t+1}= f(s_{\text{obj}}^{t}, o_{\text{tool}}^{*}, s_{\text{tool}}^{t}), \label{eq:contact_dyn}\\
        \quad & g(s^t_{\text{tool}}, s^t_{\text{obj}}) \leq 0, \quad c(s^t_{\text{tool}}, r) \leq 0.
    \end{align}
\label{eq:manip-plan}
\end{subequations}

\vspace{-0.3cm}

\noindent This formulation extends Eq.~\eqref{eq:main} by enforcing that the trajectory passes through the best-found configurations~$s_{\text{tool}}^{*}, s_{\text{obj}}^{*}$ at $T_k$. By focusing the optimization on the selected tool~$o^*_{\text{tool}}$, we avoid evaluating contact dynamics and robustness for all $N$ candidates, reducing computational complexity.

\subsection{Optimizer}
To solve the keyframe optimization problem in Eq.~\eqref{eq:keyframe_opt}, we employ Covariance Matrix Adaptation Evolution Strategy (\textit{CMA-ES})\cite{hansen2016cma}. We normalize all continuous optimization variables to the range $[0, 1]$. Discrete tool selection is handled by partitioning this range into $N$ segments, each corresponding to a specific tool, enabling joint optimization of both tool identity and continuous tool-object configurations.
For the full trajectory optimization problem described in Eq.~\eqref{eq:manip-plan}, we utilize the Via-Point-Based Stochastic Trajectory Optimization (\textit{VPSTO}) algorithm~\cite{jankowski2023vp}, a variant of CMA-ES. \textit{VPSTO} introduces via-points as a low-dimensional, time-continuous representation of trajectories, which facilitates efficient optimization across the full planning horizon in complex, high-dimensional input spaces.

\section{Learning Energy-Informed Robustness Guidance}\label{sec:metric_learning}
To practically estimate the robustness guidance~$Q$ in Eq.~\eqref{eq:main}, we adopt a simple data-driven method to learn the energy-based metric~$Q_{\text{mee}}$ and the clearance-based metric~$Q_{\text{pcc}}$, as detailed below.

\subsection{Data Collection}
We curate a dataset of $N_s$ samples for each object-tool pair by randomizing their configurations within task-relevant bounds. For each sampled configuration, we use the BIT* algorithm~\cite{gammell2015batch} to approximate the underlying Minimum Escape Energy (MEE) in a physical simulator, which serves as our robustness metric under gravitational, elastic, or pushing force fields. 
For example, in the scissor hanging task with hooks (Fig.~\ref{fig:overview}, top-right), we randomize the poses of the scissors~$s_{\text{obj}}$ for each hook. If it is collision-free, we compute the scissors' escape trajectory and corresponding robustness value~${Q}_{\text{mee}}$ using BIT*. If an escape path cannot be found (${Q}_{\text{mee}}=\inf$), we consider the scissors completely caged and define a binary cage status label $q_{\text{c}}=1$. Otherwise, the scissors are either energy-bounded caged or not caged, with ${q_{\text{c}}=0}$. The resulting dataset consists of tuples ${(s_{\text{obj}}, s_{\text{tool}}, q_{\text{c}}, {Q}_{\text{mee}})}$, thus capturing diverse configurations and their robustness metrics, enabling effective supervised learning of manipulation robustness. Similarly, a dataset ${(s_{\text{obj}}, s_{\text{tool}}, q_{\text{pc}}, {Q}_{\text{pcc}})}$ is created for the baseline robustness metric.

\subsection{Network Structure}
We employ a Multi-Layer Perceptron (MLP) to learn the robustness metric in a supervised manner. Specifically, the input consists of the object's and the tool's configurations~$s_{\text{obj}}$, $s_{\text{tool}}$. The network comprises three fully-connected hidden layers with dimensions of 128, 128, and 64 neurons, respectively, each followed by ReLU activation functions and batch normalization. Furthermore, the output layer simultaneously predicts the cage status~$\hat{q}_{\text{c}}$ via a sigmoid-activated neuron and the MEE~$\hat{Q}_{\text{mee}}$ through a linear regression neuron. Training is performed using a combined loss of binary cross-entropy for classification and Smooth L1 loss for regression, with the latter computed only on samples with $q_{\text{c}} = 0$ (i.e., not completely caged).
In the manipulation planner (Section~\ref{sec:planner}), we define a unified robustness metric~$\mathring{Q}_{\text{mee}}$ as:
\begin{equation}
\mathring{Q}_{\text{mee}} = \mathds{1}[\hat{q}_{\text{c}} \geq q_{\text{thres}}] \cdot Q_{\text{max}} + \mathds{1}[\hat{q}_{\text{c}} < q_{\text{thres}}] \cdot \hat{Q}_{\text{mee}},
\label{eq:unified_mee}
\end{equation}
\noindent where $\mathds{1}[\cdot]$ denotes the indicator function, $q_{\text{thres}} \in [0, 1]$ is a cage status confidence threshold, and $Q_{\text{max}}>0$ is a predefined high robustness value.  Eq.~\ref{eq:unified_mee} ensures assigning high robustness~$Q_{\text{max}}$ only when the object is confidently caged (above threshold). Otherwise, it is considered an energy-bounded caging configuration, relying on the learned continuous metric~$\hat{Q}_{\text{mee}}$.

For the PCC baseline robustness metric, we employ the same network structure, which outputs partial caging status~$\hat{q}_{\text{pc}}$ and clearance~$\hat{Q}_{\text{pcc}}$. These predictions are integrated in the planner through a unified partial caging clearance metric~$\mathring{Q}_{\text{pcc}}$:
\begin{equation}
\mathring{Q}_{\text{pcc}} = \mathds{1}[\hat{q}_{\text{pc}} \geq q^\prime_{\text{thres}}] \cdot (Q^\prime_{\text{max}}+\hat{Q}_{\text{pcc}}),
\end{equation}
where $q^\prime_{\text{thres}} \in [0, 1]$ is the partial caging confidence threshold and $Q^\prime_{\text{max}} > 0$ is a predefined clearance offset constant.

\section{Experiments}
We evaluate our method on three representative tool-use tasks—{tape pulling}, {fish scooping}, and {scissors hanging}—each involving selection from a set of $N {=} 3$ candidate tools. These tasks span rigid, articulated and deformable objects, covering both non-prehensile and prehensile manipulation scenarios. Experiments on the scissors hanging task validate the real-world practicality of our method.

\begin{table*}[t]
\centering
\scriptsize
\caption{VLM confidence scores (mean$\pm$std) for tool selection.}
\label{tab:vlm_results}
\begin{tabular}{lccc}
\toprule
Task & Tool 1 & Tool 2 & Tool 3 \\
\midrule
Tape pulling & 0.13$\pm$0.03 (mugtree) & 0.41$\pm$0.15 (cloth hanger) & \textbf{0.46$\pm$0.17} (umbrella) \\
Fish scooping & 0.22$\pm$0.06 (wide shovel) & 0.12$\pm$0.04 (fish slice) & \textbf{0.66$\pm$0.03} (shovel) \\
Scissors hanging & 0.18$\pm$0.07 (coat hook) & \textbf{0.64$\pm$0.01} (treble hook) & 0.18$\pm$0.06 (slatwall hook) \\
\bottomrule
\end{tabular}
\end{table*}

\ifthenelse{\boolean{includeFigures}}{
\begin{figure}[t]
    \centering
    \includegraphics[width=\linewidth]{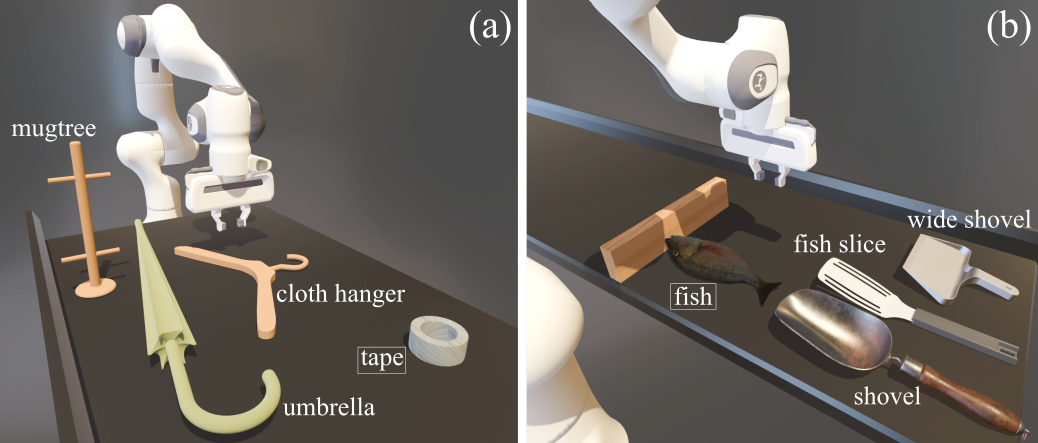}
    \caption{Our planner identifies the most robust tool from a cluttered table and plans an optimal trajectory to robustly pull a tape~(a) or scoop a fish~(b).}
    \label{fig:exp_setups}
\end{figure}
}{}

\subsection{Task and Method Description}

We evaluate our method on three representative tool-use tasks (Fig.~\ref{fig:exp_setups}, \ref{fig:rw}):  
(i) \textbf{Tape Pulling:} A Franka Panda robot pulls a Tape Roll ($s_{\text{obj}} \in SE(2)$) into a reachable region using one of three tools—Cloth Hanger, Umbrella, or Mugtree.  
(ii) \textbf{Fish Scooping:} A deformable Fish ($s_{\text{obj}} \in SE(3) \times \mathbb{R}^9$), modeled as 10 links connected by 9 compliant revolute joints,  is scooped from a surface near a fixed block using a Fish Slice, Wide Shovel, or Shovel.  
(iii) \textbf{Scissors Hanging:} A pair of articulated Scissors ($s_{\text{obj}} \in SE(3) \times \mathbb{R}$) is hung onto a Coat Hook, Treble Hook, or Slatwall Hook.  

We compare four methods: \texttt{MEE} (our proposed learned escape-energy metric), \texttt{PCC} (clearance-based baseline), \texttt{VLM} (Vision–Language Model), and \texttt{NoRob} (planning without any robustness guidance, $\beta=0$).

\subsubsection{VLM-based baseline} 
We choose {VLMs} as a baseline for tool selection because pre-trained foundation models have shown strong capabilities in scene understanding and decision-making for robotic manipulation tasks. Recent works demonstrate that VLMs can generate robotic tool designs~\cite{gao2025vlmgineer}. Our work extends the idea to tool selection using VLMs.
We use GPT-5 across the three tasks. For each task, the input image (Fig.~\ref{fig:exp_setups}a, \ref{fig:exp_setups}b, \ref{fig:rw}a) includes geometry, size, and pose information of three candidate tools, the target object, and the robot arm. The input text contains the task description, robustness definition, and other contextual details. We instruct the VLMs to reason step-by-step and output: the chosen tool, confidence scores for each tool, the reasoning process, and a sketch of the low-level action plan using the selected tool. More details can be found in the \href{https://sites.google.com/view/robust-tool-use/home}{project website}\footnote{\url{https://sites.google.com/view/robust-tool-use/}}.

\subsection{Robustness-Aware Tool Selection}
\subsubsection{How tool geometry affects robustness}
We evaluate whether the learned robustness metric $\mathring{Q}_{\text{mee}}$ produces meaningful tool–object configurations by analyzing their resistance to disturbances. For each tool, the tool–object configuration is optimized using CMA-ES (20 iterations, 100 candidates/iteration) with \texttt{MEE}, and the resulting disturbed object positions are visualized in Fig.~\ref{fig:robust_tool_selection}. Across all tasks, tools with stronger caging affordances yield higher robustness: in Tape Pulling, Mugtree fully secures Tape Roll by enclosing it in its core, Umbrella provides partial enclosure, and Cloth Hanger offers minimal constraint; in Fish Scooping, Shovel holds Fish more securely than Fish Slice or Wide Shovel; in Scissors Hanging, disturbance tests show scissors typically fall from Coat Hook at around 18\,N, from Slatwall Hook at 30\,N, and remain on Treble Hook up to 40\,N, confirming Treble Hook as the most robust choice. These results demonstrate that tools offer varying levels of robustness and that robustness guidance enables the planner to reliably identify configurations that maximize disturbance resistance for a given tool.

\subsubsection{Tool selection across methods}
We compare tool selection outcomes from \texttt{MEE}, \texttt{PCC}, \texttt{NoRob}, and \texttt{VLM} under robot kinematic constraints. Results for \texttt{MEE}, \texttt{PCC}, and \texttt{NoRob} are averaged over 20 keyframe optimization runs, while \texttt{VLM} results (Table~\ref{tab:vlm_results}) are averaged over 5 runs.

In Tape Pulling, all methods select Umbrella as it is the only tool satisfying the reachability constraint, although Fig.~\ref{fig:robust_tool_selection} shows that Mugtree can provide the most robust configuration by fully caging Tape Roll through its core. Compared to \texttt{PCC} and \texttt{NoRob}, \texttt{MEE} produces more robust Umbrella–Tape Roll keyframe configurations. \texttt{VLM} fails to identify the most robust configuration for Umbrella or Mugtree, reflecting limited capability to repurpose everyday objects for novel tool-use strategies, despite showing some zero-shot ability to infer simple geometric properties.

In Fish Scooping, \texttt{MEE} and \texttt{VLM} consistently select Shovel, which, as confirmed earlier, provides the highest disturbance resistance. In contrast, \texttt{PCC} and \texttt{NoRob} vary in their selections and often choose less robust tools and suboptimal keyframe configurations.

\ifthenelse{\boolean{includeFigures}}{
\begin{figure}[t]
    \centering
    \begin{subfigure}{\linewidth}
        \centering
        \includegraphics[width=\linewidth]{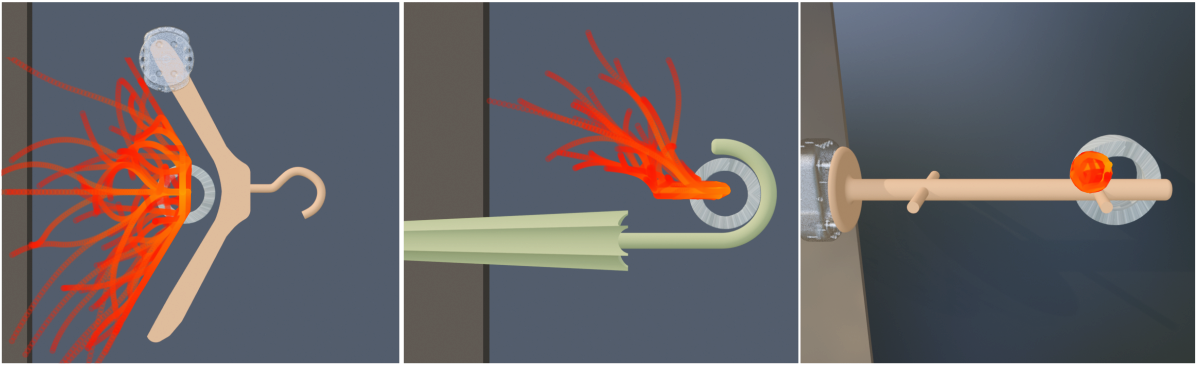}
        \caption{tape pulling}
        \vspace{0.1cm}
        \label{fig:tape_pull_tool_select}
    \end{subfigure}
    \begin{subfigure}{\linewidth}
        \centering
        \includegraphics[width=\linewidth]{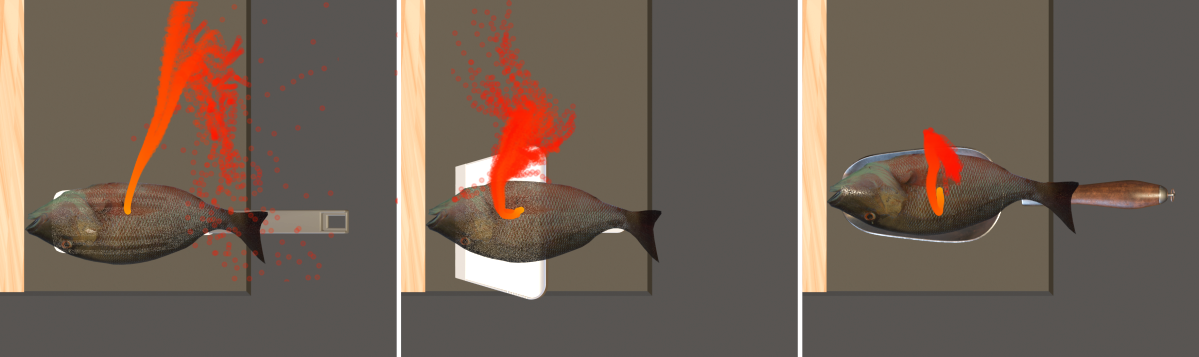}
        \caption{fish scooping}
        \vspace{0.1cm}
        \label{fig:fish_scoop_tool_select}
    \end{subfigure}
    \caption{Robustness of different tools under disturbances, evaluated in their best-found configurations~($s^*_{\text{obj}}$, $s^*_{\text{tool}}$) with \texttt{MEE}. For each tool, 100 simulated episodes of object trajectories under random force disturbances are shown in red. Tools are ordered from left to right by increasing manipulation robustness to external disturbances.}
    \label{fig:robust_tool_selection}
\end{figure}
}{}

\subsection{Robustness-Aware Tool-Use Manipulation Planning}
We assess the effectiveness of our method \texttt{MEE} for robustness-aware manipulation planning by comparing it with the baselines \texttt{PCC} and \texttt{NoRob}. The evaluation measures positional deviations of the manipulated objects under random force disturbances in open-loop execution. Results are summarized in Fig.~\ref{fig:exp_trajs} and Table~\ref{tab:fish_scoop_robustness}.

Fig.~\ref{fig:exp_trajs} shows that plans generated with \texttt{MEE} exhibit greater robustness against disturbances than those from the baselines. In Tape Pulling, all methods select Umbrella as the optimal tool due to kinematic constraints. However, \texttt{MEE} produces keyframe configurations and trajectories in which the Umbrella’s handle maintains close, consistent contact with Tape Roll (Fig.~\ref{subfig:tape_pulling_mp_ours}). This results in higher escape energy along the trajectory, making Tape Roll more resistant to external forces. In contrast, \texttt{PCC} fails to find any partial caging configurations between Umbrella and Tape Roll. As a result, its trajectories start from arbitrary collision-free pulling configurations. Lacking robustness guidance, the robot pulls Tape Roll with Umbrella using excessive force, causing it to slide out of the handle (Fig.~\ref{subfig:tape_pulling_mp_bl}) and making the system less robust to external disturbances.

In Fish Scooping, \texttt{MEE} enables Shovel to capture a substantial portion of Fish, align with it, and push it deep into Shovel’s belly using the table fixture (Fig.~\ref{subfig:fish_scoop_mp_ours}). The resulting keyframe configuration effectively constrains Fish between Shovel and the table fixture. Without effective robustness guidance, \texttt{PCC} fails to achieve this outcome (Fig.~\ref{subfig:fish_scoop_mp_bl}), producing configurations that are far more susceptible to external disturbances.

Quantitative results in Table~\ref{tab:fish_scoop_robustness} confirm the performance gap: \texttt{MEE} consistently yields higher robustness than \texttt{PCC}, while \texttt{NoRob} suffers from substantial positional deviations due to the absence of robustness guidance. These findings validate that incorporating robustness metrics into manipulation planning can significantly improve the disturbance resistance of tool-use strategies.

\ifthenelse{\boolean{includeFigures}}{
\begin{figure}[t]
    \centering
    \begin{subfigure}{1\linewidth}
        \centering
        \includegraphics[width=\linewidth, trim=0.0cm 0.5cm 0.0cm 0.0cm, clip]{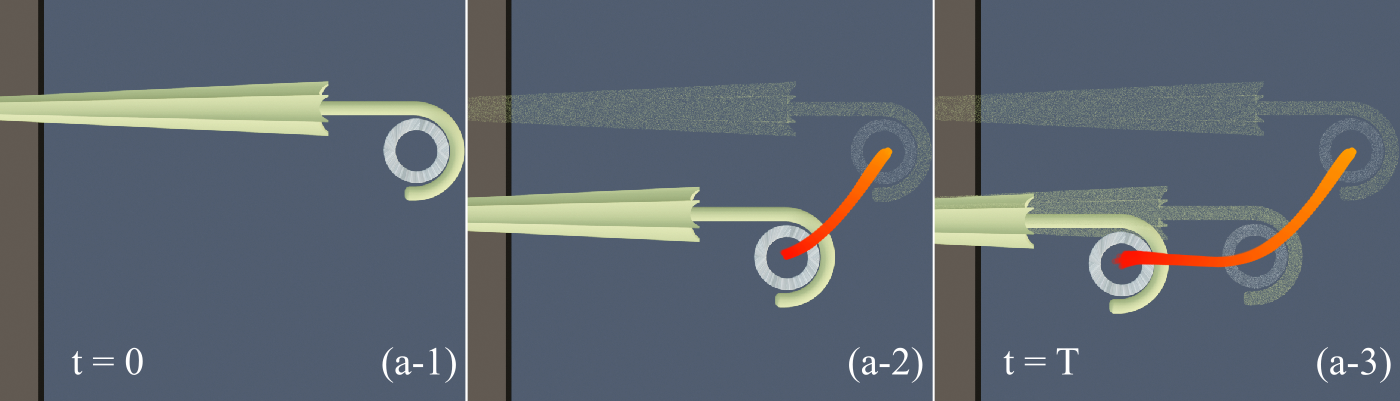}
        \caption{tape pulling (\texttt{MEE}, ours)}
        \vspace{0.1cm}
        \label{subfig:tape_pulling_mp_ours}
    \end{subfigure}
    \begin{subfigure}{1\linewidth}
        \centering
        \includegraphics[width=\linewidth, trim=0.0cm 0.5cm 0.0cm 0.0cm, clip]{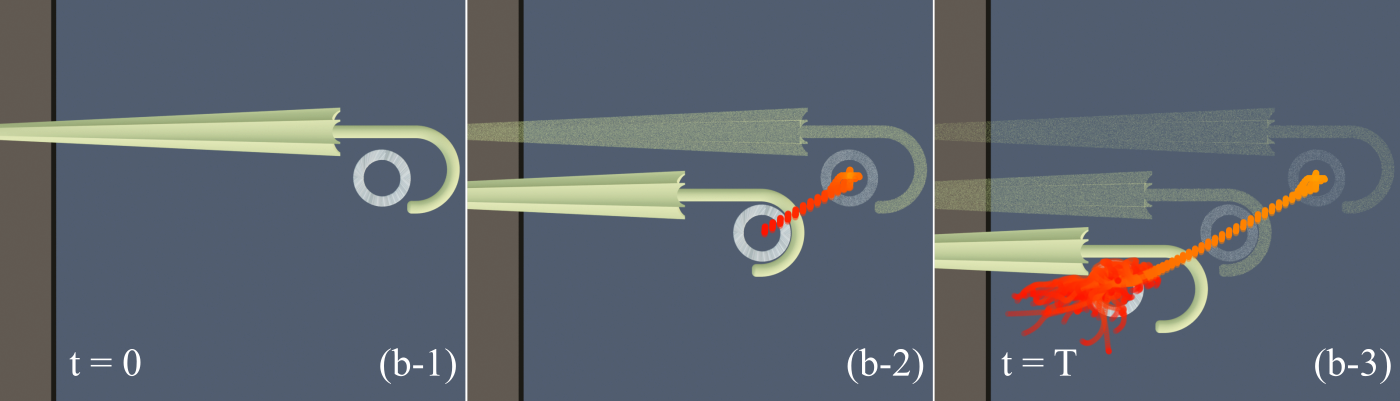}
        \caption{tape pulling (\texttt{PCC}, baseline)}
        \vspace{0.1cm}
        \label{subfig:tape_pulling_mp_bl}
    \end{subfigure}
    \begin{subfigure}{1\linewidth}
        \centering
        \includegraphics[width=\linewidth, trim=0.0cm 0.5cm 0.0cm 0.0cm, clip]{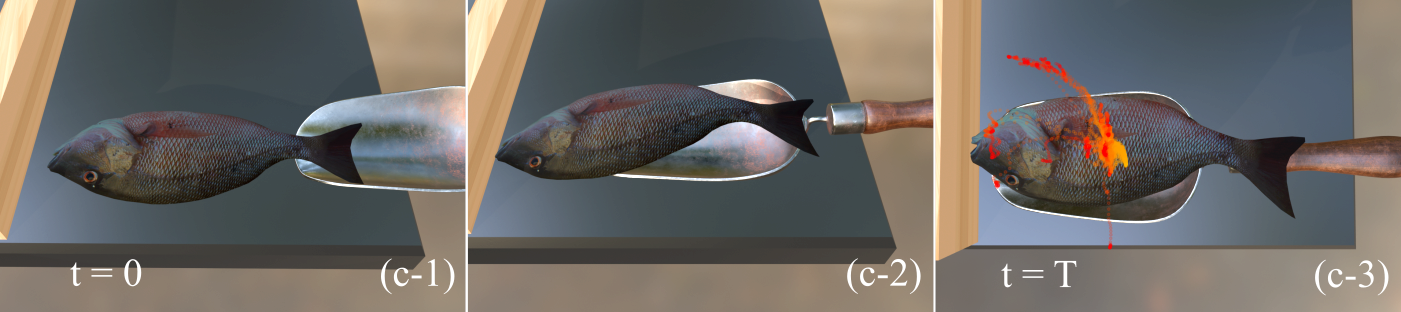}
        \caption{fish scooping (\texttt{MEE}, ours)}
        \vspace{0.1cm}
        \label{subfig:fish_scoop_mp_ours}
    \end{subfigure}
    \begin{subfigure}{1\linewidth}
        \centering
        \includegraphics[width=\linewidth, trim=0.0cm 0.5cm 0.0cm 0.0cm, clip]{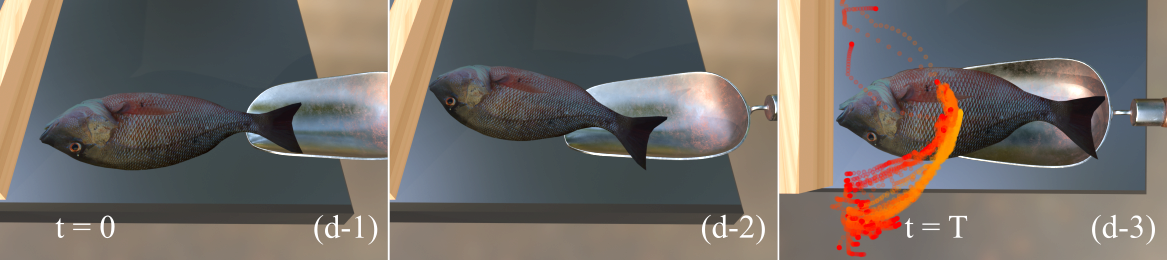}
        \caption{fish scooping (\texttt{PCC}, baseline)}
        \vspace{0.1cm}
        \label{subfig:fish_scoop_mp_bl}
    \end{subfigure}
    \caption{Manipulation trajectories planned with our method \texttt{MEE} compared to the baseline \texttt{PCC} in Tape Pulling and Fish Scooping. Snapshots show initial ($t{=}0$, left), intermediate (middle), and final ($t{=}T$, right) configurations. Under random force disturbances, the intermediate and final Tape Roll/Fish configurations from \texttt{MEE} exhibit greater disturbance resistance than those from \texttt{PCC} (red trajectories).}
    \label{fig:exp_trajs}
\end{figure}
}{}

\begin{table}[h]
\centering
\scriptsize
\begin{tabular}{lcccc}
\toprule
{Tasks} & \texttt{MEE} (Ours) & \texttt{PCC} & \texttt{NoRob} \\
\midrule
Tape Pulling & $\bm{0.054 \pm 0.021}$ & $0.237 \pm 0.136$ & $0.261 \pm 0.136$ \\
Fish Scooping & $\bm{0.190 \pm 0.267}$ & $0.761\pm0.649$ & $1.463\pm0.322$ \\
\bottomrule
\end{tabular}
\vspace{5pt}
\caption{Average positional deviations of target object under random force disturbances in open-loop execution of planned manipulation trajectory. Lower values indicate higher robustness.}
\label{tab:fish_scoop_robustness}
\end{table}

\subsection{Real World Experiments} 
We evaluate the practical applicability of our method in Scissors Hanging. The real-world experimental setup replicates the simulation environment, utilizing a UFactory xArm 7-DoF robotic arm equipped with a suction-cup gripper and an in-hand camera. In each trial, the scissors are randomly placed on the workspace, after which the robot estimates their pose and grasps them by suction. The robot then executes the planned Cartesian trajectory, transporting the scissors to the designated pre-release pose above the best-found tool~$o^*_{\text{tool}}$, i.e. {Treble Hook}. The suction is subsequently disengaged, and Scissors drop onto the hook.

Table~\ref{tab:real-world} showcases preliminary results of our method's transferability to the real world. The results indicate the benefit of explicitly optimizing for robustness by maximizing escape energy during keyframe selection.
Failure cases reveal that the primary reasons for unsuccessful attempts were possibly errors in pose estimation and slight physical misalignments. More examples of failure, such as misaligned suction positions or collisions between the scissors and the hooks or wall along the executed trajectory, can be found on our \href{https://sites.google.com/view/robust-tool-use/home}{website}. We also evaluated in the real-world experiments the most promising baseline method \texttt{PCC} from the simulation results (over \texttt{VLM} and \texttt{NoRob}). It is more prone to uncertainties, since the partial caging metrics fail to effectively characterize manipulation robustness in the presence of physical contacts and gravitational potential energy. The results demonstrate the practical utility of our proposed robustness metric and its potential in addressing real-world manipulation uncertainties.

\begin{table}[h]
\centering
\scriptsize
\begin{tabular}{lcc}
\toprule
\textbf{Method} & \texttt{MEE} & \texttt{PCC} \\
\midrule
SR & 0.83 & 0.50 \\
\bottomrule
\end{tabular}
\vspace{5pt}
\caption{Success rates in real-world Scissors Hanging executions under uncertainties. 20 runs in total.}
\label{tab:real-world}
\end{table}

\ifthenelse{\boolean{includeFigures}}{
\begin{figure}[t]
    \centering
    \includegraphics[width=1.0\linewidth]{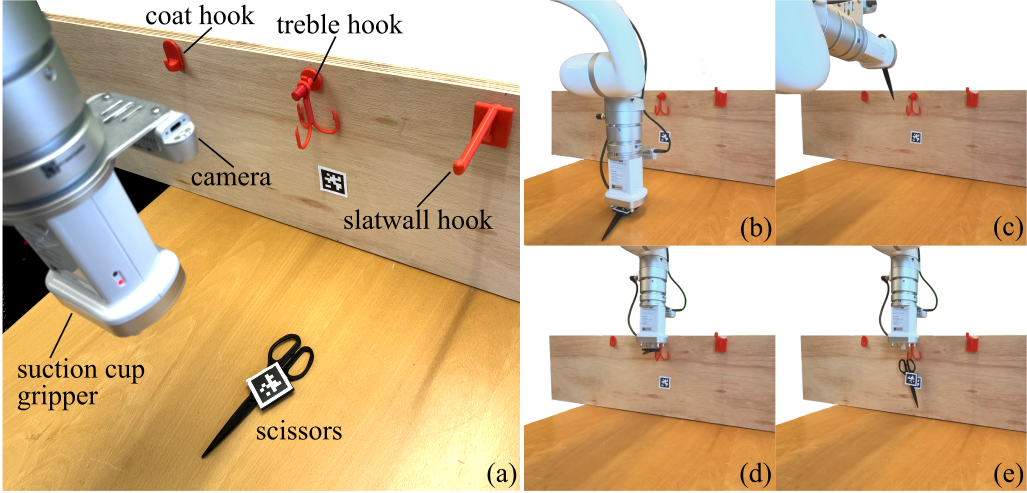}
    \caption{
    Real-world experiment setup of the scissors hanging task~(a) with a successful manipulation trajectory (b-e). The robot picks up the scissors on the table and places them on the selected \textit{Treble Hook} in the most robust way.
    }
    \label{fig:rw}
\end{figure}
}{}

\subsection{Implementation Details} 
For robustness guidance learning, the number of training samples~$N_s$ per object–tool pair ranges from $1000$ to $5000$, depending on the dimensionality of the target object's configuration space and the complexity of its collision geometry. Key hyperparameters include learning rate of $4\!\times\!10^{-5}$, batch size of $32$, $200$ training epochs, and the Adam optimizer with a weight decay of $1\!\times\!10^{-4}$. All models are trained on an Intel i9-12900H CPU. We publicly release the robustness guidance dataset for our three tasks on our \href{https://sites.google.com/view/robust-tool-use/home}{website}. It comprises tool–object configurations with corresponding MEE/PCC values, along with tool and object model files.

For manipulation planning, we use VP-STO for full-horizon trajectory optimization with $5$ via-points, $100$ candidates per iteration, and $50$ iterations. In {Tape Pulling}, planning involves (i) reaching the keyframe configuration~$(s^*_{\text{tool}}, s^*_{\text{obj}})$ while keeping the tape stationary, followed by (ii) executing a robustness-aware manipulation trajectory~$\tau^*_{\text{tool}}$ guided by the robustness metric. In {Fish Scooping}, the shovel is first positioned under the fish’s tail, after which the robot scoops the fish toward the keyframe configuration. 

During evaluation, random force disturbances are applied at each of the $100$ execution steps, with magnitudes matching the object’s weight. In Tape Pulling, disturbances are applied at four equidistant points (top, bottom, left, and right) along the outer circumference of Tape Roll, directed radially outward from the center toward the contact point. In Fish Scooping, similar planar disturbances are applied at the center of mass of Fish in the $x$–$y$ plane. Additionally, a randomized gravity-compensation force, ranging from $50\%$ to $75\%$ of the fish’s weight, is introduced to assess the robustness of the generated trajectories.

\section{Discussion}
We leverage a simple neural network to predict the robustness of tool-object configurations, significantly improving computational efficiency. Each configuration requires approximately 3 seconds to evaluate using BIT*, making direct evaluation during planning prohibitively expensive. Instead, our learned robustness module enables fast forward inference on a CPU with an average time of $0.057 \pm 0.019$\,ms over $1000$ trials. To plan a full open-loop trajectory, our hierarchical planner queries the robustness model approximately $50 \times 100 \times 100 = 5 \times 10^5$ times, resulting in a total inference time of about 29 seconds---orders of magnitude faster than the $\sim$17 CPU-days that would be needed without the neural network. 
This efficiency enables practical real-time planning while maintaining strong regression accuracy, as shown in Fig.~\ref{fig:violin}.

One limitation of our method is that the robustness metric is trained only on the tools and objects present in the evaluated tasks. Future work will extend the dataset to include a broader variety of tools and objects, enabling a network with zero-shot generalization capabilities to novel tools and objects. In particular, a large-scale, data-driven model that maps raw camera inputs directly to robustness values would be especially useful. With such a metric, the proposed planner could be applied more broadly for tool selection and manipulation planning under disturbances. On the other hand, our method currently leverages random force disturbances as a proxy for hard-to-model system dynamic mismatches. A promising direction for future work is to integrate belief-space planning approaches~\cite{daniels2023grasping, jankowski2024planning} that explicitly account for robustness against system uncertainties.

\ifthenelse{\boolean{includeFigures}}{
\begin{figure}[t]
    \centering
    \includegraphics[width=0.8\linewidth]{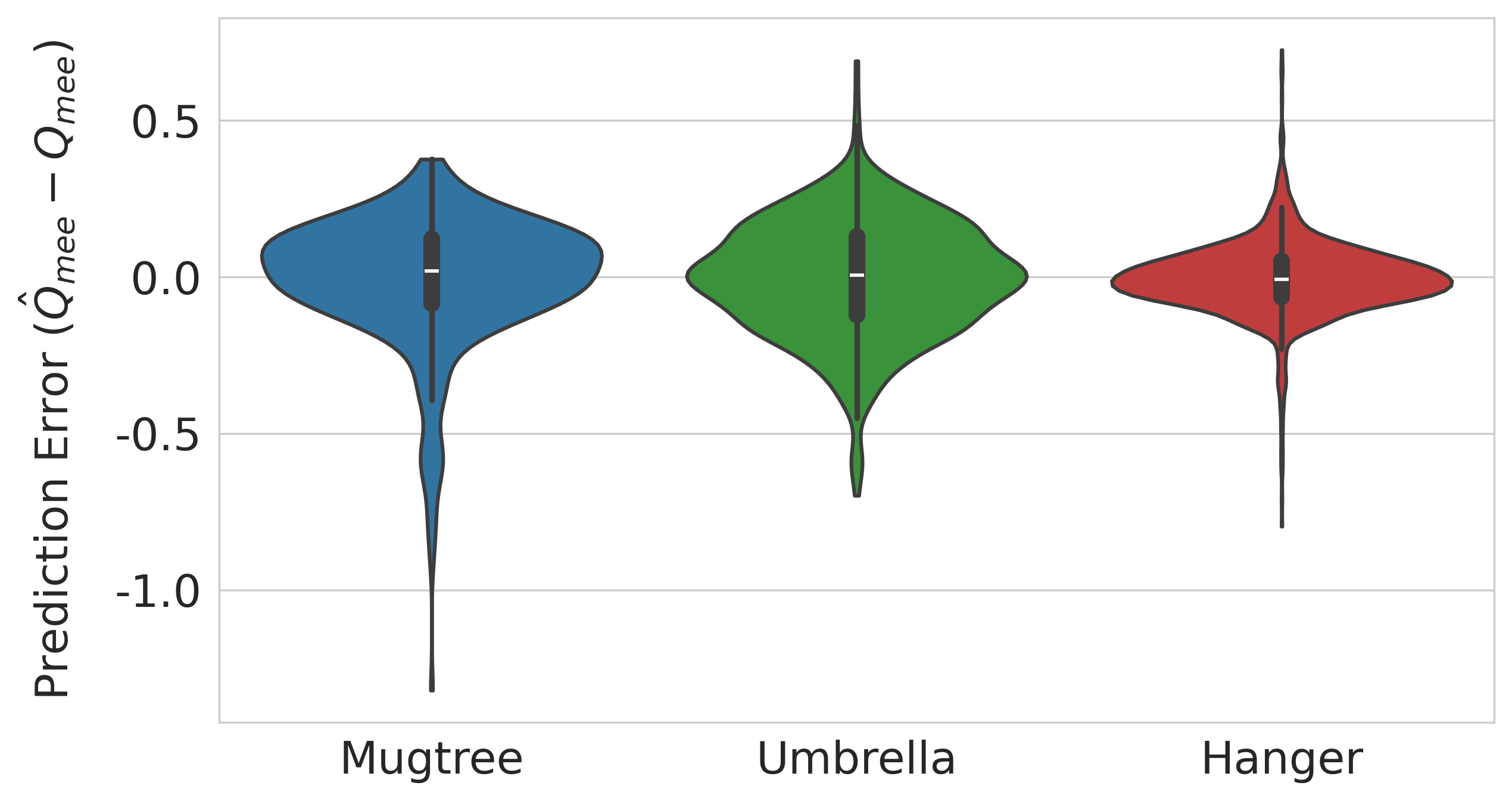}
    \caption{Prediction error of the learned robustness guidance across three object-tool pairs in the tape pulling task. The violin plot shows the distribution of absolute errors between predicted $\hat{Q}_{\text{mee}}$ and BIT*-computed ${Q}_{\text{mee}}$. White dashes: medians; black bars: interquartile ranges (IQR); whiskers: extrema within 1.5 IQR.}
    \label{fig:violin}
\end{figure}
}{}

\section{Conclusion}
Tool geometry significantly influences robustness in object manipulation. Building on this observation, we proposed a hierarchical optimization method that jointly selects a tool and plans its corresponding manipulation trajectory. By integrating an energy-informed robustness metric into the planning loop, our method consistently improved performance across three representative tasks, enabling robust manipulation under disturbances.

\bibliographystyle{bibliography/IEEEtran}
\bibliography{bibliography/IEEEabrv, bibliography/references}

\end{document}